\documentclass[numsec,webpdf,modern,large]{oup-authoring-template}
\graphicspath{{Fig/}}
\usepackage{dsfont}
\usepackage{hhline}
\usepackage{booktabs}
\usepackage{natbib}
\setcitestyle{numbers}
\setcitestyle{square}

\usepackage[mathlines, switch]{lineno}
\theoremstyle{thmstyleone}%
\theoremstyle{thmstyletwo}%
\theoremstyle{thmstylethree}%

\begin{document}

\journaltitle{Bioinformatics}
\DOI{DOI HERE}
\copyrightyear{2025}
\pubyear{2014}
\access{Advance Access Publication Date: }
\appnotes{Paper}

\firstpage{1}

\subtitle{Subject Section}

\title[SSRL on Gene Expression Data]{Self-supervised Representation Learning on Gene Expression Data}

\author[1,2,$\ast$]{Kevin Dradjat}
\author[1]{Massinissa Hamidi}
\author[2]{Pierre Bartet}
\author[1]{Blaise Hanczar}

\authormark{Dradjat et al.}

\address[1]{\orgdiv{IBISC Laboratory}, \orgname{University Paris-Saclay (Univ. Evry)}, \orgaddress{\country{France}}}
\address[2]{\orgname{ADLIN}, \orgaddress{\postcode{Paris}, \country{France}}}

\corresp[$\ast$]{Corresponding author. \href{email:email-id.com}{kevin.dradjat@univ-evry.fr}}

\received{Date}{0}{Year}
\revised{Date}{0}{Year}
\accepted{Date}{0}{Year}

%\editor{Associate Editor: Name}

\abstract{
\textbf{Motivation:} Predicting phenotypes from gene expression data is a crucial task in biomedical research, enabling insights into disease mechanisms, drug responses, and personalized medicine. Traditional machine learning and deep learning rely on supervised learning, which requires large quantities of labeled data that are costly and time-consuming to obtain in the case of gene expression data. Self-supervised learning has recently emerged as a promising approach to overcome these limitations by extracting information directly from the structure of unlabeled data.\\
\textbf{Results:} In this study, we investigate the application of state-of-the-art self-supervised learning methods to bulk gene expression data for phenotype prediction. We selected three self-supervised methods, based on different approaches, to assess their ability to exploit the inherent structure of the data and to generate qualitative representations which can be used for downstream predictive tasks. By using several publicly available gene expression datasets, we demonstrate how the selected methods can effectively capture complex information and improve phenotype prediction accuracy. The results obtained show that self-supervised learning methods can outperform traditional supervised models besides offering significant advantage by reducing the dependency on annotated data. We provide a comprehensive analysis of the performance of each method by highlighting their strengths and limitations. We also provide recommendations for using these methods depending on the case under study. Finally, we outline future research directions to enhance the application of self-supervised learning in the field of gene expression data analysis. This study is the first work that deals with bulk RNA-Seq data and self-supervised learning.\\
\textbf{Availability:} The code and results are available at \href{https://github.com/kdradjat/SSRL_RNAseq}{https://github.com/kdradjat/ssrl-rnaseq}.\\
\textbf{Contact:} \href{kevin.dradjat@univ-evry.fr}{kevin.dradjat@univ-evry.fr}\\
\textbf{Supplementary information:} Supplementary data are available at \textit{Bioinformatics}
online.}

%\keywords{Deep Learning, Self-supervised Learning, Gene Expression Data}

% \boxedtext{
% \begin{itemize}
% \item Key boxed text here.
% \item Key boxed text here.
% \item Key boxed text here.
% \end{itemize}}

\maketitle

\section{Introduction}

High-throughput sequencing technologies have made it possible to generate large amounts of transcriptomic data, offering a wide range of materials for scientific research\cite{ngs}. However, the preprocessing and analysis of these complex data presents significant challenges, particularly when it comes to building accurate and generalizable predictive models. In this article, we focus on predicting phenotypes from gene expression data. Phenotype prediction is key in biomedical research, aiding in understanding diseases, finding biomarkers, and developing personalized therapies.\\
Classical supervised machine learning methods achieve the best performance in this domain\cite{MLomics} and recent advances have shown that deep learning methods have great potential to outperform classical machine learning methods\cite{DLomics, hanczar_assessment_2022}, as deep neural networks can capture non-linear relationships between variables, in exchange of a large training dataset. These methods rely heavily on annotated data, which is costly to obtain. Besides, the available gene expression datasets do not contain as many examples as in other domains, like images or natural language processing where datasets are composed of millions of examples. This leads to a higher risk of overfitting for deep learning models and a limited generalization of the trained models on new datasets. 
\begin{figure}[!t]
    \centering
    \includegraphics[scale=0.12]{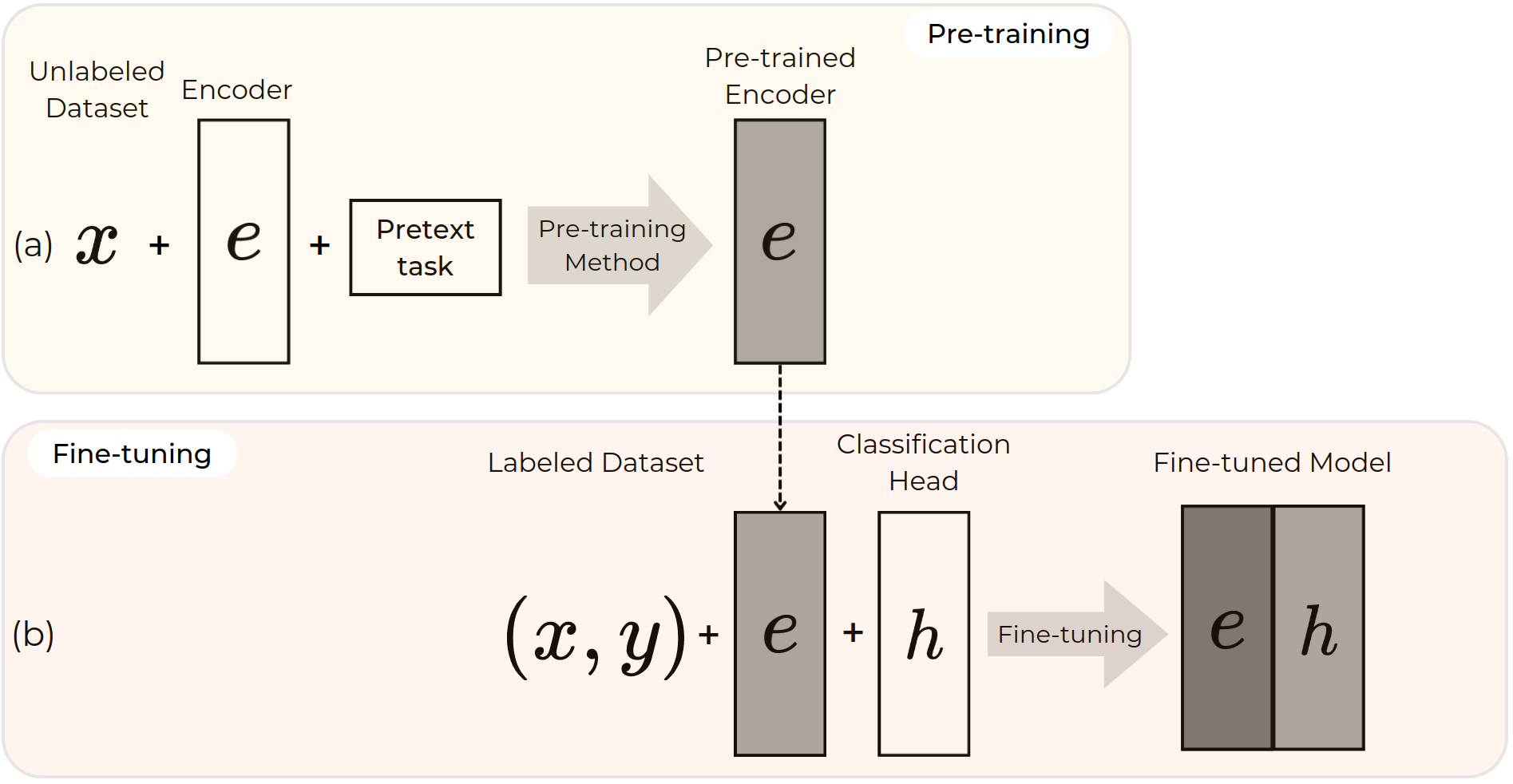}
    \caption{Self-supervised pipeline. Considering an encoder $e$, the first step consists in pre-training the encoder with unlabeled data (a). Once the pre-training is done, a classification head is attached to the encoder to train the network on a specific downstream task with labeled data (b). This second step is called fine-tuning.}
    \label{fig_ssrl}
\end{figure}
\textcolor{black}{To overcome these challenges, transfer learning and semi-supervised learning approaches are used to deal with small labeled dataset and their reliability has already been studied for gene expression data\cite{hanczar_assessment_2022, ged_semi}.}
\textcolor{black}{More recently, self-supervised learning approaches have emerged as a promising approach\cite{ssrl_review, ssrl_review_2024} and has already been shown to be efficient for related biological data such as single-cell data\cite{ssrl_sc} or medical images\cite{ssrl_medical_images}.} Indeed, in contrast to supervised learning methods, self-supervised learning does not rely on annotated data and extracts information directly from the structure of the data to establish meaningful representations that can be applied to the resolution of various downstream tasks. Self-supervised learning enables the utilization of non-annotated datasets unrelated to phenotype prediction, addressing the scarcity of annotated data and improving model performance.
%\textcolor{black}{To overcome these challenges, self-supervised learning has recently emerged as a promising approach\cite{ssrl_review, ssrl_review_2024} and has already been shown to be efficient for related biological data such as single-cell data\cite{ssrl_sc} or medical images\cite{ssrl_medical_images}}. Indeed, in contrast to supervised learning methods, self-supervised learning does not rely on annotated data and extracts information directly from the structure of the data to establish meaningful representations that can be applied to the resolution of various downstream tasks. Self-supervised learning enables the utilization of non-annotated datasets unrelated to phenotype prediction, addressing the scarcity of annotated data and improving model performance.

%\textcolor{black}{Note that alternative strategies such as transfer learning or semi-supervised learning can also be used when dealing with small labeled dataset and their reliability has already been studied for gene expression data\cite{hanczar_assessment_2022, ged_semi}.}

In this study, we propose investigating the performance of state-of-the-art self-supervised learning methods for phenotype prediction from gene expression data. We selected three self-supervised methods from different self-supervised learning categories that were initially created and applied to tabular data and evaluated their ability to generate effective representations for predictive models. Through large scale and systematic experiments using public transcriptomic datasets, we demonstrate the effectiveness of these approaches in terms of predictive accuracy and highlight their limitations. We also provide recommendations for using these methods depending on the case under study. 

By providing a comparative analysis of the performance of each selected method, this paper aims to highlight the potential of self-supervised learning to improve phenotype prediction while reducing the dependence on annotated data, offering new perspectives for biomedical research and personalized medicine.

% \begin{figure*}[!t]
%     \centering
%     \includegraphics[scale=0.14]{figures/ssrl_scheme.PNG}
%     \caption{Self-supervised pipeline. Considering an encoder $e$, the first step consists in pre-training the encoder with unlabeled data (a). Once the pre-training is done, a classification head is attached to the encoder to train the network on a specific downstream task with labeled data (b). This second step is called fine-tuning.}
%     \label{fig_ssrl}
% \end{figure*}

\section{Methods}

\subsection{Self-supervised learning}

Recent breakthroughs in supervised learning have enabled the development of high-performance models to solve various machine learning tasks. However, supervised learning requires access to a large amount of annotated data, which can be costly and time-consuming in some cases. The availability of large amounts of unlabeled data has led to major advances in different fields, such as natural language processing or the vision domain. These advances are largely due to recent self-supervised methods, which construct a new data representation space \textcolor{black}{with a pre-training step}, making prediction tasks easier \cite{ssrl_review}.
%The availability of large amounts of unlabeled data has led to major advances in different fields, such as natural language processing or vision domain, thanks to recent self-supervised methods consisting of constructing a new data representation space \textcolor{black}{with a pre-training step}, making the prediction tasks easier \cite{ssrl_review}. 

\textcolor{black}{
Formally, we consider a set of labeled data $\mathcal{D}_{l}=\{ x_{i}, y_{i} \}_{i=1}^{|\mathcal{D}_{l}|}$ of $|\mathcal{D}_{l}|$ examples, where $x_{i} \in \mathcal{X} \subset \mathbb{R}^{N}$ denotes an example and $y_{i} \in \mathcal{Y}$ its associated label. In the context of supervised learning, we aim to learn a function $f: \mathcal{X} \rightarrow \mathcal{Y}$ by minimizing a supervised loss function (e.g. MSE or cross-entropy). With the self-supervised approach, we leverage an unlabeled dataset $\mathcal{D}_{u}=\{ x_{i} \}^{|\mathcal{D}_{u}|}_{i=1}$ to learn an encoder function $e: \mathcal{X} \rightarrow \mathcal{Z}$, where $\mathcal{Z}$ is the latent space defined by a pretext task..
%where $\mathcal{Z}$ corresponds to the latent space of representations defined by a chosen pretext task. 
This first step is called pre-training. Generally, $|\mathcal{D}_{u}| \gg |\mathcal{D}_{l}|$ since unlabeled data is easier to collect than labeled data. Then, the learned representation space is used to solve various downstream tasks by using a small labeled dataset and attaching a classification head to the pre-trained encoder. This second step is called fine-tuning. The self-supervised training process is described in Figure \ref{fig_ssrl}. This approach allows a generalization, allowing the creation of robust and polyvalent models, called foundation models, which can be adapted to various downstream tasks with small annotated datasets. Self-supervised learning is mainly characterized by the strategy adopted to define the features to be predicted during the pre-training, the so-called pretext tasks. The choice of the pretext task is essential, as it impacts the representations computed by the encoder. The definition of these pretext
tasks mainly relies on the exploitation of underlying domain-specific structure of the data e.g., spatial relations for images or semantic relations for the language. Therefore, the pretext task is tightly linked to the nature of the considered data. 
}\\
\indent In this study, we focus on self-supervised methods for tabular data, as gene expression data are arranged in the form of high-dimension tabular data, where rows correspond to patients and columns correspond to genes. Besides, the internal structure of tabular data is the most challenging to discover as there are no explicit relationships between variables. \textcolor{black}{The specificity of gene expression data is the high dimensionality ($>50$k features) and low sample size. In addition to their tabular format, some genes are highly similar and therefore are expressed similarly, which can introduce redundancy in the data.}

\indent In the literature on self-supervised learning, few works has been done for tabular data\cite{ssrl_review_tab} and no methods were designed or applied on bulk RNA-seq data. Various approaches have been proposed to advance the exploitation of implicit tabular structures without labels. We distinguish three main types of approach:
\begin{itemize}
    \item Predictive/Generative Learning: This approach uses a predictive or generative task as a pretext task. The challenge lies on designing predictive tasks that are effective considering the downstream task on which the model will be applied. There is no consensus on the choice of the predictive task and several methods have been proposed eg. reconstructing examples from masked features\cite{pred_ref1}, by applying perturbations in latent space\cite{pred_ref2} or by adopting natural language processing methods\cite{pred_ref3}.
    \item Contrastive Learning: This approach aims to extract information from inter-sample similarities and differences, ensuring that two similar examples also have similar representations in latent space and inversely \cite{contr_ref1, contr_ref2, scarf}. The difficulty lies in identifying how the examples are being put together or separated, i.e choosing the adapted loss function. The current \textcolor{black}{popular approach} is to use the InfoNCE loss introduced by \cite{infonce}. 
    \item Hybrid Learning: This approach consists in using parts of predictive and/or contrastive learning or combines both. It tends to integrate the advantages of both strategies \cite{saint}. 
\end{itemize}

\subsection{Evaluated approaches methods}
%We selected three of the most popular and efficient self-supervised learning methods designed for tabular data. \textcolor{black}{We selected a method for each type of self-supervised learning approach}. 
\textcolor{black}{We chose three widely used and effective self-supervised learning methods tailored for tabular data, each representing one of the main categories of self-supervised learning: contrastive, generative, and hybrid approaches.}
The first one is SCARF \cite{scarf}, an adaptation for tabular data of SimCLR \cite{simclr}, which is a well-known self-supervised learning method used in computer vision. The second method is VIME \cite{vime}, a modified version of a denoising auto-encoder \cite{dae} designed for tabular data. The last is BYOL \cite{byol}, which is a method initially designed for computer vision based on the principle of teacher-student network. %Since most of state-of-the-art hybrid self-supervised learning methods are based on Transformers and the dimension of transcriptomic data is high, we did not select a hybrid learning method.

\begin{figure*}
    \centering
    \includegraphics[scale=0.4]{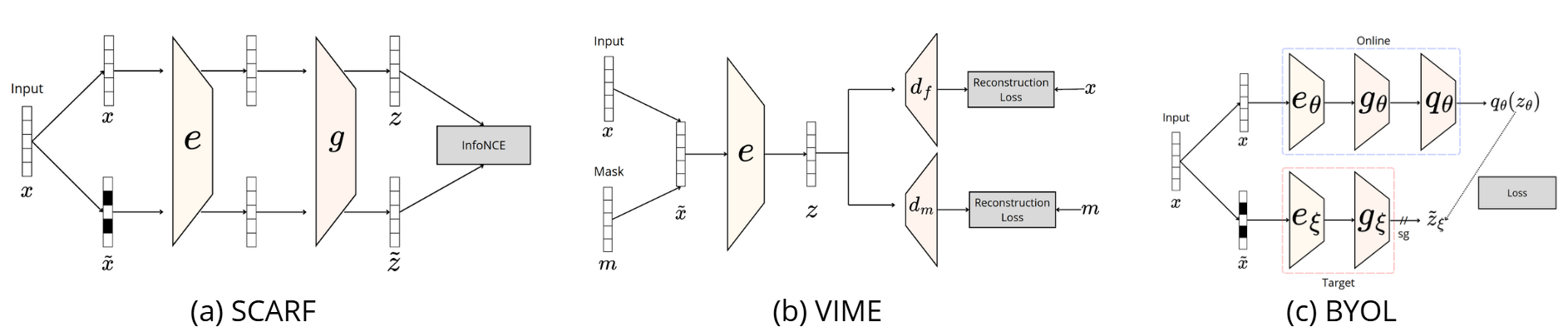}
    \caption{Architectures of the considered SSL pre-training approaches: (a) SCARF, (b) VIME, (c) BYOL}
    \label{methods}
\end{figure*}

\subsubsection{SCARF}
SCARF\cite{scarf} is a contrastive self-supervised learning method for tabular data, described in Figure \ref{methods}a. First, considering a batch of examples of the unlabeled training data, a corrupted version $\tilde{x}^{(i)}$ is generated for each example $x^{(i)}$ \textcolor{black}{by randomly selecting a fraction of its features. Each selected feature is replaced by a random draw from the continuous uniform distribution whose support is defined by the minimum and maximum values of that feature in the training set} 
%uniform distribution over the values that feature takes on across the training dataset. 
Then, the pre-training process' objective is to train an encoder $e$ that will define the representation space. The original example $x^{(i)}$ and the corrupted version $\tilde{x}^{(i)}$ pass through the encoder network $e$ to produce representations $z^{(i)}$ and $\tilde{z}^{(i)}$ that will pass through a projector $g$ to obtain projections $q^{(i)}$ and $\tilde{q}^{(i)}$ respectively. Finally, the InfoNCE contrastive loss\cite{infonce} is applied to ensure that projections $q^{(i)}$ and $\tilde{q}^{(i)}$ are close, for all $i$, and projections $q^{(i)}$ and $\tilde{q}^{(j)}$ are far apart, for $i \neq j$: 
\begin{equation}
    \color{black} \mathcal{L} = \dfrac{1}{N} \sum\limits_{i=1}^{N} -\log \left( \dfrac{\exp(s_{i,i}/\tau)}{\textcolor{black}{\sum_{\substack{k=1}}^{N}} \exp(s_{i,k}/\tau)} \right) 
\end{equation}
where $s_{i,j}=q^{(i)^\intercal} \tilde{q}^{(j)} / (\lVert q^{(i)}\rVert_{2} . \textcolor{black}{\lVert \tilde{q}^{(j)} \rVert_{2}} )$, for $i,j \in [N]$ and $\tau$ is the temperature hyperparameter.
% \begin{figure}[H]
%     \centering
%     \includegraphics[scale=0.10]{figures/scarf_process.png}
%     \caption{SCARF pre-training process.}
%     \label{fig_scarf}
% \end{figure}
After the pre-training, the encoder network $e$ is kept and attached to a classification head $h$ that takes the outputs of $e$ as its inputs.% Then, the fine-tuning part consists in optimizing the cross-entropy classification loss.

\subsubsection{VIME}
VIME\cite{vime} is a self-supervised generative auto-encoder-based learning method for tabular data that relies on two pretext tasks: feature reconstruction and mask reconstruction. First, the pre-training process \textcolor{black}{consists of} generating a binary mask vector $m = [m_{1}, \cdot, m_{d} ]^{\perp} \in \{0,1\}^{d}$ where $m_{j} \sim \mathcal{B}(p_{m})$, to generate a corrupted version $\tilde{x}$ for each example $x$ with the following transformation: $\tilde{x} = m \odot \overline{x} + (1 - m) \odot x $,
% \begin{equation}
%  \tilde{x} = m \odot \overline{x} + (1 - m) \odot x 
% \end{equation}
where the features of $\overline{x}$ are sampled from the empirical marginal distribution of each features.\\
The corrupted sample $\tilde{x}$ passes through an encoder $e$ that produces a representation $\tilde{z}$, from which a decoder $d_{f}$ predicts the values of the features that have been corrupted and a decoder $d_{m}$ predicts which features have been masked. The total loss is a combination of a feature reconstruction loss $\mathcal{L}_{f}$ and a mask reconstruction loss $\mathcal{L}_{m}$: $\mathcal{L} = \mathcal{L}_{m} + \alpha\color{black}\mathcal{L}_{f}  $, 
% \begin{equation}
%  \mathcal{L} = \mathcal{L}_{m} + \alpha\mathcal{L}_{r}    
% \end{equation}
\textcolor{black}{where $\mathcal{L}_{f}$ is the mean squared error loss, $\mathcal{L}_{m}$ is the sum of the binary cross-entropy losses for each dimension of the mask vector} and $\alpha$ is a hyper-parameter that adjusts the trade-off between the two losses.
The pre-training process is described in Figure \ref{methods}b. After the pre-training process, the encoder part $e$ is conserved and can be fine-tuned.

% \begin{figure}[H]
%     \centering
%     \includegraphics[scale=0.10]{figures/vime_process.png}
%     \caption{VIME pre-training process.}
%     \label{fig_vime}
% \end{figure}

\subsubsection{BYOL}
BYOL\cite{byol} (Bootstrap Your Own Latent) is a self-supervised learning method for visual data. We have made some modifications to this method to adapt it to tabular data. BYOL is a pre-training method composed of two neural networks branches: \textit{online} and \textit{target}. The \textit{online} branch is defined by a set of weight $\theta$ and composed of three parts: an encoder $e_{\theta}$, a projector $g_{\theta}$ and a predictor $q_{\theta}$. The \textit{target} branch is defined by a set of weight $\xi$ and is composed of an encoder $e_{\xi}$ and a projector $g_{\xi}$. \\
% \begin{figure}[H]
%     \centering
%     \includegraphics[scale=0.10]{figures/byol_process.png}
%     \caption{BYOL pre-training process.}
%     \label{fig_byol}
% \end{figure}
For the pre-training process, we choose to generate a corrupted version $\tilde{x}$ of an example $x$ in the same way as VIME. Then, the \textit{online} branch takes as input the original example $x$ to produce the \textcolor{black}{prediction $q_{\theta}(g_{\theta}(e_\theta(x)))=q_{\theta}(z_{\theta})$}. Similarly, the \textit{target} branch takes as input the corrupted examples $\tilde{x}$ to produce the \textcolor{black}{projection $g_{\xi}(e_{\xi}(\tilde{x}))=\tilde{z}_{\xi}$}. 
The $l2-$distance between the \textcolor{black}{outputs} is then minimized with the following $\mathcal{L}_{\theta, \xi}$ loss: 
%The loss $\mathcal{L}_{\theta,\xi}$, defined as follows, is applied: 
\begin{equation}
\color{black}
  \mathcal{L}_{\theta, \xi} = \lVert \overline{q_{\theta}(z_{\theta})} - \overline{\tilde{z}_{\xi}} \rVert_{2}^{2} = 2 - 2 \cdot \dfrac{\langle q_{\theta}(z_{\theta}), \tilde{z}_{\xi} \rangle}{\lVert q_{\theta}(z_{\theta}) \rVert_{2} \cdot \lVert \tilde{z}_{\xi} \rVert_{2}}    
\end{equation}
\textcolor{black}{We use an overline to denote normalized quantities.}
\textcolor{black}{BYOL is considered as a hybrid approach because it learns from positive pairs only.} During pre-training, at each training step, the loss $\mathcal{L}_{\theta, \xi}$ is minimized with respect to $\theta$ only, as the \textit{target} branch is on \textit{stop-gradient} mode. On the other hand, the weights $\xi$ are updated by doing an exponential moving average of the online parameters $\theta$. More precisely, at each training step and given a target decay rate $\lambda \in [0,1]$, the following update is applied: $\xi \leftarrow \lambda \xi + (1 - \lambda) \theta$.
% \begin{equation}
%      
% \end{equation}
 The pre-training process is described in Figure \ref{methods}c. After the pre-training, only the encoder $e_{\theta}$ is conserved for the fine-tuning process.\\

\textcolor{black}{
\textbf{Effects of data augmentation.}
The objective of applying transformations to the examples before the training process is to diversify the training features. Although these transformations do not have biological meaning, they simulate biological diversity well while preserving the label. 
Indeed, one risk here is that the transformations alter the examples so much that their corresponding labels change.
To support this statement, we conducted an experiment which can be found in Supplementary Material (Figure S5). We assume that they do not fundamentally change the underlying information for two main reasons.
    First, since we consider thousands of features and we perturb only a fraction of them ($30\%$), the transformations' effect is limited.
    Then, because of redundancy, mentioned in Section 2.1, the applied transformations have less effect on the examples, since the information contained in a given feature can be contained in another feature.
    }

\section{Experiments and Results}

\subsection{Datasets}\label{dataset}
We selected two RNA-Seq datasets, described in detail in Supplementary with the preprocessing method (Section A), to conduct our experiments: 
\begin{itemize}
    \item The Cancer Genome Atlas (TCGA): this is a pan-cancer RNA-Seq dataset that contains 9349 samples and 56902 input genes from 19 different types of cancer, \textcolor{black}{with the majority class representing 12.9\% of the dataset and the smallest class representing 2.8\%.}
    \item All RNA-Seq and ChIP-Seq Sample and Signature Search (ARCHS4)\cite{archs4}: this is a pan-tissue samples originating from experiences from SRA (Sequence Read Archive) and GEO (Gene Expression Omnibus). It is a dataset that has not yet been widely used in the field of deep learning. It is composed of diverse samples originating from experiences that do not only focus on cancer. It therefore represents a good candidate for use as pre-training dataset. 
    %We have filtered the samples to create a dataset adapted to our experiments, 
    It contains 53282 samples and 67128 input genes from 19 different tissue types, \textcolor{black}{with the majority class representing 14.1\% of the dataset and the smallest class representing 0.6\%.} 
\end{itemize}

%\textbf{Preprocessing.} To preprocess the ARCHS4 dataset, we first extract the different tissue types with the ARCHS4py Python package. We then normalize samples and correct for differences in gene count distribution before applying the batch-effect correction tool pyComBat\cite{pycombat}, following the process described on the ARCHS4 platform. We applied the same normalization process to the TCGA dataset. The preprocessing was performed before splitting the datasets into pre-training and fine-tuning sets.

\subsection{Experimental Setup}
\subsubsection{Evaluation settings} \label{evaluation}

The goal is to evaluate the effectiveness of the three selected self-supervised methods with the TCGA and ARCHS4 RNA-Seq datasets.
Although self-supervised methods use unlabeled data for the pre-training process, both datasets are labeled. Thus, we divided each dataset into an unlabeled pre-training and a labeled fine-tuning set while conserving initial class proportions to simulate using unlabeled examples during the pre-training. We define three evaluation settings depending on the dataset considered. In each case, we pre-train the model with the unlabeled pre-training set and fine-tune with it with the labeled fine-tuning set on the downstream task, consisting in cancer type identification (multi-class classification):
\begin{itemize}
    \item TCGA: We divided the dataset into a pre-training set composed of 7291 examples and a fine-tuning set and test set both composed of 1029 examples. 
    \item ARCHS4: We divided the dataset into a pre-training set containing 50998 examples and a fine-tuning set and test set both containing 1142 examples. 
    \item ARCHS4 to TCGA: We pre-train the model with the pre-training ARCHS4 dataset and fine-tune it with the fine-tuning TCGA dataset. The goal is to see if the representation learned from the ARCHS4 samples can be transferred to the TCGA downstream task. 
    As the TCGA and ARCHS4 datasets do not feature the same set of genes, we only consider the common genes between the two datasets, representing 55747 genes.
\end{itemize}
In each case, we take $15\%$ of both the pre-training and fine-tuning set as validation sets. \textcolor{black}{According to \ref{dataset}, a majority class classifier would achieve 12.9\% of accuracy on the TCGA and ARCHS4-to-TCGA case, and 14.1\% on the ARCHS4 case.}\\

\textbf{Model architecture and training.} We tested many encoder architectures $e$ by varying the number and size of hidden layers. \textcolor{black}{We selected the best architecture by training on the fine-tuning training set and taking the one with the best accuracy on the fine-tuning validation set. The hyper-parameters tested are listed in Supplementary (Table S4)}. The best encoder architecture is composed of four dense layers of dimension 256 and with ReLU activation function and batch normalization. \\%, following \cite{scarf}. \\
For SCARF, the pre-training head $g$ is a 2-layers ReLU network with hidden dimension 256. For VIME, we use the same architecture as the encoder $e$ for both the feature decoder $d_{f}$ and mask decoder $d_{m}$ with an output dimensionality of the same size of the input dimensionality. The mean square error reconstruction loss is applied for the feature decoder $d_{f}$ and the binary cross-entropy loss is applied for the mask decoder $d_{m}$. For BYOL, inspired by \cite{byol}, we choose a linear layer with output size 4096 followed by batch normalization, ReLU and a final linear layer with output dimension 256 as projectors $g_{\theta}$ and $g_{\xi}$ and predictor $q_{\theta}$. The pre-training hyper-parameters are listed in Supplementary (Table S3). \textcolor{black}{We use the pre-training validation set for loss tracking during the pre-training step.}

% \begin{table}[!h]
%     \centering
%     \begin{tabular}{c  c  c  c}
%         \toprule
%         & SCARF & VIME & BYOL\\
%         \midrule
%         pre-training epochs & 1000 & 500 & 50 \\
%         \midrule
%         batch size & 256 & 32 & 32 \\
%         \midrule 
%         optimizer & Adam & RMSprop & SGD \\
%         \midrule
%         learning rate & $10^{-4}$ & $10^{-3}$ & $10^{-4}$ \\
%         \midrule 
%         momentum & $\times$ & $\times$ & 0.9\\
%         \midrule
%         corruption rate $c$ & 0.3 & 0.3 & 0.3 \\
%         \midrule 
%         temperature $\tau$ & 1.0 & $\times$ & $\times$ \\
%         \midrule 
%         trade-off $\alpha$ & $\times$ & 2.0 & $\times$ \\
%         \midrule 
%         decay rate $\lambda$ & $\times$ & $\times$ & 0.9 \\
%         \bottomrule
%     \end{tabular}
%     \caption{Hyper-parameters for each methods.}
%     \label{tab_data}
% \end{table}

For the fine-tuning part, we associate the pre-trained encoder with a classification head composed of one linear layer with embedding size as input dimension and the number of classes as output dimension. We consider two cases: the unfrozen fine-tuning where the layers of the encoder and the classification are trained, and the frozen fine-tuning where we freeze the layers of the pre-trained encoder and only the parameters of the classification head are trained. %Supervised fine-tuning uses early stopping \textcolor{black}{using the loss on fine-tuning validation set}.\\ 
Supervised fine-tuning uses early stopping with patience 30, \textcolor{black}{using the loss on fine-tuning validation set}, with a max number of epochs of 100 and a batch size of 8.\\ 

\textbf{Benchmarking.} We apply each pre-training method on the defined encoder and compare their performances, after fine-tuning, to the performances of its non-pre-trained version. We call the non-pre-trained model the baseline model. For the fine-tuning part, the training set comprises approximately 1000 examples. We perform fine-tuning with different proportions $p$ of examples used during training, from 0.02 to 1 with a step of 0.01, to highlight the effect of fine-tuning with very few examples.
 To estimate the variability of the approaches, we train 5 models for each proportion and each method by randomly initializing the weights of the fine-tuning head in the pre-trained case and by randomly initializing the weights of the whole network for the baseline model. 
\begin{figure*}
    \includegraphics[scale=0.4]{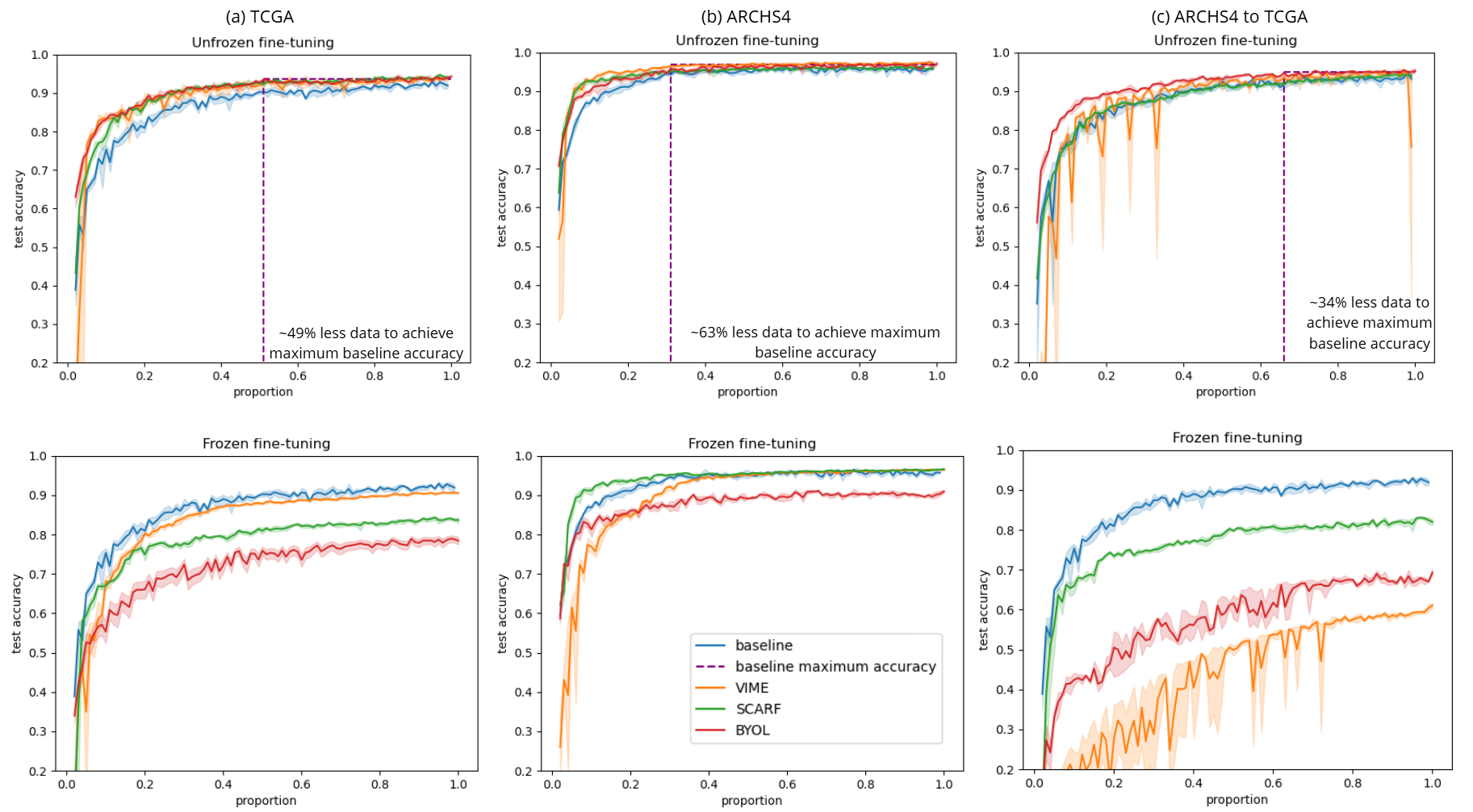}
    \caption{Performance of pre-trained models compared to non-pre-trained model (baseline) for the three different cases: (left column) TCGA, (center column) ARCHS4 and (right column) ARCHS4 to TCGA.}
    \label{acc_total}
\end{figure*}

% \begin{figure*}[!t]
% \centering
% \begin{tabular}{@{}c@{}}
%    \includegraphics[scale=0.27]{figures/acc_tcga.PNG} \\[\abovecaptionskip]
%    \small (a) TCGA: Accuracy on TCGA dataset with encoders pre-trained on TCGA pre-training set.
% \end{tabular}
% \begin{tabular}{@{}c@{}}
%    \includegraphics[scale=0.27]{figures/acc_archs4.PNG} \\[\abovecaptionskip]
%    \small (b) ARCHS4: Accuracy on ARCHS4 dataset with encoders pre-trained on ARCHS4 pre-training set.
% \end{tabular}
% \begin{tabular}{@{}c@{}}
%    \includegraphics[scale=0.27]{figures/acc_transfer.PNG} \\[\abovecaptionskip]
%    \small (c) ARCHS4 to TCGA: Accuracy on TCGA dataset with encoders pre-trained on ARCHS4 pre-training set.
% \end{tabular}
% \caption{Performance of pre-trained models compared to non-pre-trained model (baseline) for the three different cases: TCGA, ARCHS4 and ARCHS4 to TCGA.}
% \label{acc}
% \end{figure*}

\paragraph{\textbf{Results of evaluation}}
The results are presented in Figure \ref{acc_total}. For the two first cases in which pre-training and fine-tuning examples originate from the same dataset (\ref{acc_total}-first row). We notice a clear difference in performance of the pre-trained models between the frozen and the unfrozen fine-tuning. \\
Indeed, in the unfrozen case, pre-trained models achieve better performance than the baseline model, especially at low fine-tuning proportion $p$ for the three experiments. This clearly shows the benefits of self-supervised approaches to improve prediction performances. \textcolor{black}{Besides, all models seem to reach the same accuracy as the proportion increases. However, they seem to reach this common point faster when considering the ARCHS4 dataset.} This may be due to the fact that in the case of ARCHS4, the models were pre-trained with more examples than in the TCGA case. In the first two experiments, the three self-supervised methods return the same performance. The third experiment, where pre-training and fine-tuning data come from different datasets, is the most challenging task. We note that BYOL strongly outperforms the other methods.  

In the frozen case, the performances are variable. BYOL always produces lower performances. VIME is generally worse than the baseline, but may reach the same performances with a high proportion $p$. These two approaches seem not to draw benefits from frozen pre-training. SCARF may be worse than the baseline or outperform the baseline, especially for low proportion $p$ (see \ref{acc_total}a,c). These results show that frozen self-supervised learning is possible but is clearly more complex than unfrozen self-supervised learning and requires methods specific to gene expression data.

%These results show that the selected pre-training methods seem to produce better models than without pre-training when the fine-tuning is carried out on unfrozen mode and when the pre-training and fine-tuning examples come from the same dataset. When we try to use the representation built using the ARCHS4 dataset on the TCGA dataset, the pre-training does not produce better models than without pre-training, at best it permits to achieve the same performance as the baseline does. In all cases, the frozen fine-tuning does not produce high-performance models. The difference in behavior of the pre-trained models between the unfrozen fine-tuning and the frozen fine-tuning demonstrates that the representations produced do not initially fit to the downstream task but can be useful when they are slightly modified i.e. by performing the fine-tuning on unfrozen mode. \\
The selected pre-training methods improve model performance when fine-tuning is done with unfrozen layers and both pre-training and fine-tuning use the same dataset. However, transferring representations from ARCHS4 to TCGA does not yield better results than training from scratch. In all cases, frozen fine-tuning leads to poor performance, highlighting that the learned representations require adaptation to the downstream task. \\
By performing the fine-tuning on unfrozen mode, we saw that pre-trained models can outperform the baseline. When this is the case, pre-trained models need less examples to reach the maximum baseline accuracy. For the TCGA case (Figure \ref{acc_total}a) and the ARCHS4 to TCGA case (Figure \ref{acc_total}c), the maximum baseline accuracy is first achieved by the model pre-trained with BYOL, saving respectively 49\% and 34\% of the TCGA training dataset. For the ARCHS4 case (Figure \ref{acc_total}b), the maximum baseline accuracy is achieved by the model pre-trained with VIME and permits to save approximately 63\% of the ARCHS4 training set. Thus, these self-supervised methods permits to produce models that can be effective and can outperform the non-pre-trained model with less examples, which is a great advantage when we consider that annotated RNASeq data is time-consuming and expensive to obtain and existing RNASeq datasets contain few examples.

\textcolor{black}{The benefits of self-supervised learning can also be demonstrated when we look at the convergence speed (evolution of accuracy as a function of training epochs) of pre-trained models during fine-tuning. Indeed, fine-tuning a pre-trained model takes less training epochs than starting from random weights. Self-supervised methods permit to build models that converges faster to a local minimum by providing a better initialization.}
%Indeed, the fine-tuning phase of the pre-trained models lasts less epochs than the non-pre-trained one: the model pre-trained with SCARF converges in 15 epochs, the one pre-trained with BYOL converges in 10 epochs and the model pre-trained with VIME in 28 epochs, whereas the training of the baseline model lasts 30 epochs. Thus, the self-supervised methods permits to build models that converges faster to a local minimum.

This observation makes us consider the pre-training step as an efficient model initialization, leading to a faster convergence and lesser data consumption in supervised training.

\begin{figure*}[!t]
    \centering
    \includegraphics[scale=0.22]{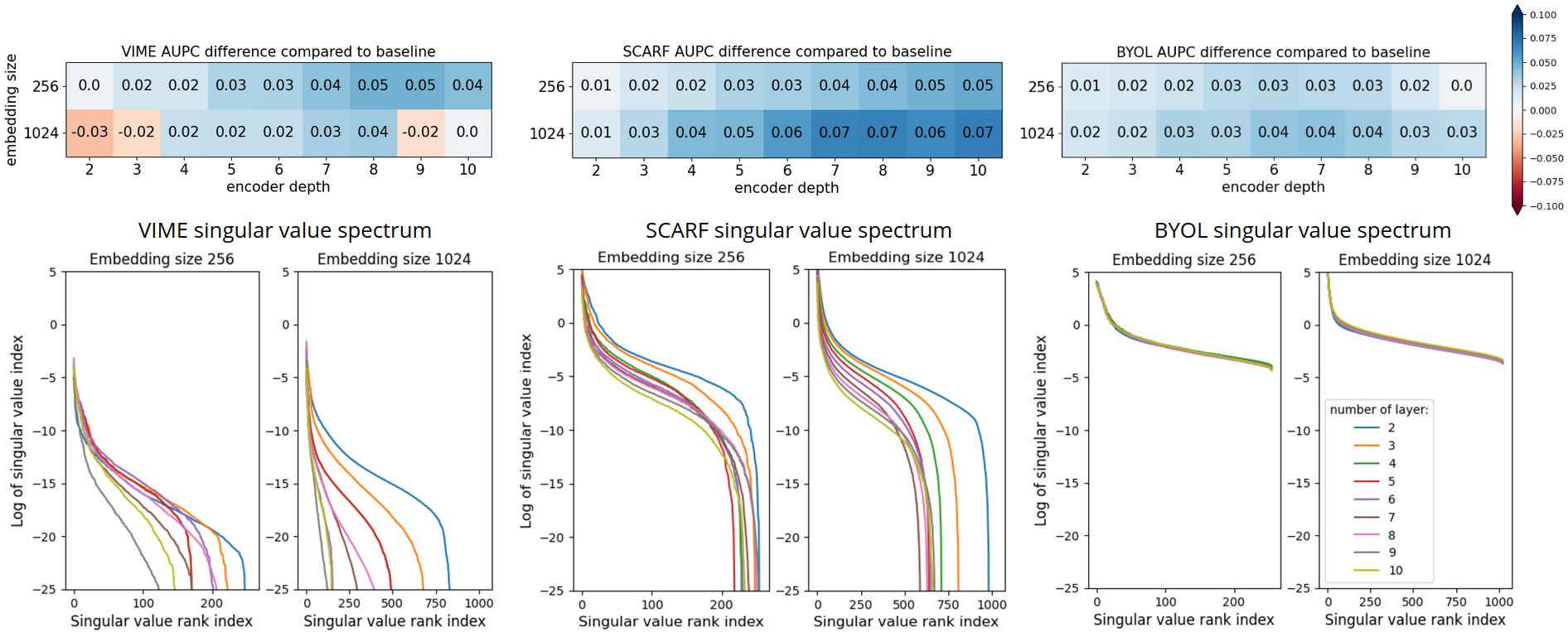}
    \caption{(Top) Average gain of different encoder architecture under the three pre-training methods compared to the associate baseline model coupled to (Down) singular value spectrum for each architecture.}
    \label{archi_loop}
\end{figure*}

\subsubsection{Variation of architecture's size} \label{archi}
We want to highlight the architecture-dependent factor of the representation induced by pre-training methods mentioned in \cite{dl_survival}. We adopt the same evaluation process as mentioned in \ref{evaluation} except that we vary the number of layers and the embedding size of the considered model. The number of layers varies from 2 to 10 and the embedding size varies in the range $[256, 1024]$. We then pre-train and fine-tune, in unfrozen mode, several architectures with each pre-training methods and compare each one with its corresponding not pre-trained version. 
%The models fine-tuning is carried out in unfrozen mode. 
To estimate the global performance gap between a pre-trained model and a its non-pre-trained version, we adopt a metric that compares the area under the performance curves (AUPC).

More precisely, we only consider the AUPC for $p$ varying from 0.02 to 0.3, representing 20 to 300 examples, corresponding to the size of a large majority of the real datasets. The corresponding metric is:
\begin{equation}
  g(m_{1}, m_{2}) = AUPC(m_{1})_{[0.02, 0.3]} - AUPC(m_{2})_{[0.02, 0.3]}   
\end{equation}
 
where $m_{1}$ and $m_{2}$ represent the accuracy curves of the two models considered. We then obtain a metric so that the model $m_{1}$ performs better on average than $m_{2}$ if $g(m_{1}, m_{2}) > 0$ and inversely if $g(m_{1}, m_{2}) < 0$.
This metric estimates the performance of a pre-trained model and allows the comparison of different methods on a range of training set size.\\

\paragraph{\textbf{Results of variation of architecture's size}}
The results are presented in Figure \ref{archi_loop}. When we fix the embedding size at 256, we can see that the three types of pre-training are beneficial regardless of the depth of the encoder. The metric value ranges from 0.00, for VIME with 2 layers, to 0.07, for SCARF with 10 layers. When we increase the embedding size to 1024, as the model is deeper, the VIME pre-training becomes less effective until reaching a negative value, showing counterproductive pre-training. We can then suppose that the VIME pre-training method can be ineffective in improving models with large embedding size and can produce highly architecture-dependent representations. Concerning SCARF and BYOL, the pre-training is still beneficial with embedding size of 1024, showing that the representations induced by this method are not architecture-dependent.

%For a more in-depth analysis, we studied the dimensional collapse of the learned representations. Dimensional collapse is defined as the shrinking of representation vectors along certain dimension, where the model does not use some dimensions of the latent space. Since this phenomenon is a sign of low-quality representation learning and can potentially affect performance on downstream tasks \cite{collapse}, we conducted a study on our pre-trained models to see the quality of the representation for each architecture. 
For a more in-depth analysis, we analyzed dimensional collapse-where the model underutilizes parts of the latent space, indicating porr representation learning and potential downstream performance issues\cite{collapse}. We examined this phenomenon across our pre-trained models for each architecture.
We analyze learned representations by collecting the embedding vectors on the validation set. Following \cite{collapse}, we evaluate the dimensionality by first collecting the embedding vectors on the validation set. We then compute the covariance matrix $C \in \mathcal{M}_{d\times d}(\mathbb{R})$ of the embedding vectors: $C = \dfrac{1}{N} \sum_{i=1}^{N}(z_{i}-\overline{z})(z_{i}-\overline{z})^{\top} $,
% \begin{equation}
%  C = \dfrac{1}{N} \sum_{i=1}^{N}(z_{i}-\overline{z})(z_{i}-\overline{z})^{\top}    
% \end{equation}
where $\overline{z} = \dfrac{1}{N}\sum_{i=1}^{N}z_{i}$ is the mean and $N$ is the total number of samples. We then apply the singular value decomposition on this matrix ($C=USV^{\top}$, where $S=diag(\sigma^{k})$) and dispose them in sorted order and logarithmic scale, as shown in Figure \ref{archi_loop}. When several singular values collapses to zero, it is a sign of dimensional collapse. Figure \ref{archi_loop} shows the singular value spectrum for each architecture.\\
We can see that models pre-trained with VIME show signs of dimensional collapse, especially with deeper architectures. We can suppose that this collapse is the cause of bad performance on downstream tasks as VIME's gains are unstable with deeper architectures. For example, with the model composed of 9 layers with an embedding size of 1024, approximately only 100 singular values do not collapse to zero and it is associated to a negative performance model in downstream task. \textcolor{black}{In contrast, SCARF and BYOL are more prone to dimensional collapse due to the design of the pre-training process based on the use of positive and/or negative pairs}. However, we can see that models pre-trained with SCARF present a less pronounced dimensional collapse: for an embedding size of 1024 up to 50\% of singular values collapse to zero and up to 20\% collapse to zero for an embedding size of 256. Those pre-trained with BYOL do not present this effect regardless of the number of layer and the embedding size. This confirms the fact that the design of BYOL pre-training is robust to dimensional collapse which is a key point of good performances in our case. Interestingly, SCARF shows a relative effect of dimensional collapse but presents good performances on downstream task. 

To conclude, this analysis shows that BYOL permits to build representations that use all the available dimensions, regardless of the architecture, whereas it is not the case for VIME and SCARF. However, dimensional collapse is not a sign of bad representation learning in our case, as 
even collapsed models can outperform baselines on downstream task.
%a model that presents collapse can be used for downstream task and can still outperform the non-pre-trained version.

\subsubsection{Variation of number of samples in pre-training set}
\begin{figure}[!h]
    \centering
    \includegraphics[scale=0.45]{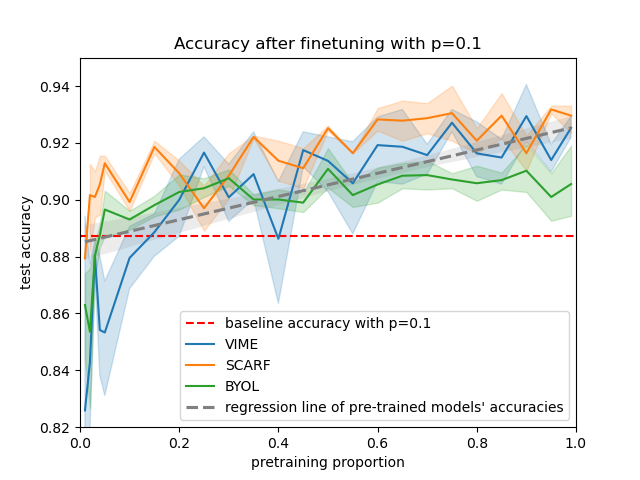}
    \caption{Accuracy of pre-trained models with different pre-training proportion and with fine-tuning proportion $p=0.1$.}
    \label{pretrain_prop}
\end{figure}

We also aim to observe the effect of the pre-training set size. For that, we use the ARCHS4 datasets with the initial architecture considered in \ref{evaluation}. We vary the number of examples used during pre-training by increasing the proportion $q$ of the pre-training set used from 0.01 to 0.05 with a step of 0.01, then from 0.05 to 1 with a step of 0.05. We train the model for each pre-training proportion $q$ used and for each pre-training method and compare its performance to the non-pre-trained version. For the comparison between two methods, we choose to consider the performance of each model with proportion $p=0.1$ used in fine-tuning. For this experiment, we set the fine-tuning part on unfrozen mode. The results are presented in Figure \ref{pretrain_prop}.

\paragraph{\textbf{Results on the pre-training set size}}
The results presented in Figure \ref{pretrain_prop} show the effect of the size of the pre-training set on the accuracy of the models with a fine-tuning proportion $p$ fixed to $0.1$. We observe that the pre-training is more effective as the pre-training proportion increases and can be already effective with a few examples in the pre-training set. More precisely, the SCARF and BYOL pre-training methods provide increasing gains from $q=0.05$ compared to the baseline, representing approximately $2500$ examples. The same goes for VIME which provides increasing gains from $q=0.2$. We can also see that pre-trained models achieve lower performances than the baseline when the pre-training proportion is small, showing that pre-training can be counterproductive when using an insufficient amount of example during this phase. Note that for BYOL and SCARF, this counterproductive effect occurs at very low pre-training proportion, representing approximately $2000$ examples. In contrast, for VIME the effect occurs when there are fewer than $10000$ examples in the pre-training set.

\section{Discussion}
%\subsection{Advantages and Limitations}
Results show that self-supervised methods can enhance neural network performance on gene expression data, but only under certain conditions. 
%The results obtained so far show that self-supervised methods can improve the performance of neural networks in the field of gene expression data. In the case of the architecture used in \ref{evaluation}, the pre-training process permits to enhance the performance of the model under specific conditions. 
Indeed, first the fine-tuning process has to be done with unfrozen layers, otherwise the performance of the pre-trained models does not outperform the baseline model performance. It shows that the representations induced by the three selected pre-training methods can not be used as they are and must be slightly modified through an unfrozen fine-tuning. Pre-training is most effective when datasets are homogeneous. If the datasets are homogeneous, then the pre-trained models outperform the baseline model, especially at low fine-tuning proportion $p$. This is a promising result concerning pre-training applied to gene expression data because a low fine-tuning proportion corresponds to a clinical application case for which only a few data are available ($<1000$ examples). If pre-training and fine-tuning data differ (e.g., due to batch effects), only the BYOL method is effective. SCARF and VIME fail to outperform the baseline in case of dataset heterogeneity even if we ensure that the two datasets are processed in the same way and include the same classes. % and that the classes of the fine-tuning set are included in the pre-training set. % biggest performance gap happens in low proportion fine-tuning that correpsonds to real life cases
BYOL is the only method that provides better performances when the pre-training and the fine-tuning set originates from different datasets, showing that the pre-training added qualitative value to the representations. The lack of improvements in this case could be due to the misalignment of the data classes. Indeed, even if the two different datasets follow the same distribution by applying the batch effect correction, those of the TCGA classes are not aligned with the distributions of the ARCHS4 classes. 

In addition, the representations induced by the pre-training methods can be highly architecture-dependant. Thus, choosing the right architecture is key to avoid counterproductive pre-training, as seen for VIME with high-layered models. However, SCARF and BYOL appear less architecture-dependent as they only improve the baseline model by improving the performance with a few examples in fine-tuning and no counterproductive cases were encountered.

Finally, the dataset used for the pre-training does not need to contain many examples for this step to be effective. Indeed, we observed improvements for models pre-trained with only a few examples $(\sim9000)$. However, for these experiments, we considered the case where the pre-training and fine-tuning examples come from the ARCHS4 dataset, which can be considered as the most straightforward case because the ARCHS4 dataset contains only 10 classes for the downstream task. Since it can be considered as an easy downstream task, a short pre-training with a few examples can easily improve the model.

\section{Conclusion}
This study explores the use of self-supervised learning methods on gene expression data for phenotype prediction. We showed self-supervised learning creates high-quality representations from unlabeled data. Self-supervised learning is a cornerstone for the construction of foundation models for gene expression that may significantly improve the performance of classical supervised learning models. The three selected methods succeeded in this task, particularly when the fine-tuning is made with very few data, which is close to a real application case. Thus, these methods reduce reliance on expensive, time-consuming labeled data in the domain of gene expression data. \\
%Besides, to tackle the fact that the effect of self-supervised pre-training is variable
Besides, to address variability in SSL effectiveness, we provide recommendations for using these methods to observe gains compared to the baseline. Indeed, the fine-tuning should be performed on unfrozen mode. Then, if the pre-training and fine-tuning sets are homogeneous, we recommend using the SCARF method because its pre-training phase lasts less time than that of BYOL. If the sets are not homogeneous, we recommend using BYOL because it is the only method that outperforms the baseline in this case. Practitioners should be cautious with VIME, as its pre-training is longer and the results in fine-tuning are very unstable compared to the two other methods.  \\
Although the results are promising, some research challenges and prospects remain, for example the batch effect correction (heterogeneity of pre-training and fine-tuning data) or the design of efficient pretext tasks. \textcolor{black}{We focus our study on bulk gene expression data, but as omics data share similarities, this work should be applied to related data (genomics, proteomics, etc). We therefore hope that this work will encourage future research in the development of self-supervised approaches specific to omics data.} 
%%%%%%%%%%%%%%

%\begin{appendices}

%\end{appendices}

\section{Competing interests}
No competing interest is declared.

\section{Author contributions statement}
All authors conceived the presented idea and discussed the results and reviewed the manuscript. K.D and P.B contributed to the design and implementation of the computations. K.D, B.H and M.H wrote the manuscript. 

\section{Funding}
This work has been supported by the Paris \^Ile-de-France R\'egion in the framework of DIM AI4IDF and by a collaboration with ADLIN Science.

\section{Acknowledgments}
The results shown here are in part based upon data generated by the TCGA Research Network, publicly available here: \url{https://www.cancer.gov/tcga}. We are grateful to the Institut Français de Bioinformatique for providing computing and storage resources.

\bibliographystyle{abbrvnat}
\bibliography{references}

%USE THE BELOW OPTIONS IN CASE YOU NEED AUTHOR YEAR FORMAT.
%\bibliographystyle{abbrvnat}
%\bibliography{reference}

\end{document}